\documentclass{article}

\usepackage{microtype}
\usepackage{graphicx}
\usepackage{subfigure}
\usepackage{booktabs} % for professional tables

\usepackage{hyperref}

\usepackage[accepted]{icml2020_arxiv}
\usepackage{enumerate}
\usepackage[fleqn]{amsmath}
\usepackage{amsthm,scalefnt}
\usepackage{amscd}
\usepackage{amssymb}

\usepackage{graphics} % for pdf, bitmapped graphics files
\usepackage{mathptmx} % assumes new font selection scheme installed
\usepackage{multirow}
\usepackage{amsbsy}
\usepackage{array}
\usepackage{calrsfs}
\usepackage{anyfontsize}

\usepackage{color}
\usepackage{amsmath,amsthm}

\DeclareMathAlphabet{\mathcal}{OMS}{cmsy}{m}{n}
\SetMathAlphabet{\mathcal}{bold}{OMS}{cmsy}{b}{n}

\icmltitlerunning{Online Passive-Aggressive Total-Error-Rate Minimization}

\begin{document}

\twocolumn[
\icmltitle{Online Passive-Aggressive Total-Error-Rate Minimization}

% It is OKAY to include author information, even for blind
% submissions: the style file will automatically remove it for you
% unless you've provided the [accepted] option to the icml2020
% package.

% List of affiliations: The first argument should be a (short)
% identifier you will use later to specify author affiliations
% Academic affiliations should list Department, University, City, Region, Country
% Industry affiliations should list Company, City, Region, Country

% You can specify symbols, otherwise they are numbered in order.
% Ideally, you should not use this facility. Affiliations will be numbered
% in order of appearance and this is the preferred way.
\icmlsetsymbol{equal}{*}

\begin{icmlauthorlist}
\icmlauthor{Se-In Jang}{to}
\end{icmlauthorlist}

\icmlaffiliation{to}{Department of Statistics and Applied Probability, National University of Singapore, Singapore}

\icmlcorrespondingauthor{Se-In Jang}{sijang@nus.edu.sg}

% You may provide any keywords that you
% find helpful for describing your paper; these are used to populate
% the "keywords" metadata in the PDF but will not be shown in the document
%\icmlkeywords{Machine Learning, ICML}

\vskip 0.3in
]

% this must go after the closing bracket ] following \twocolumn[ ...

% This command actually creates the footnote in the first column
% listing the affiliations and the copyright notice.
% The command takes one argument, which is text to display at the start of the footnote.
% The \icmlEqualContribution command is standard text for equal contribution.
% Remove it (just {}) if you do not need this facility.

\printAffiliationsAndNotice{}  % leave blank if no need to mention equal contribution
%\printAffiliationsAndNotice{\icmlEqualContribution} % otherwise use the standard text.

\begin{abstract}
	We provide a new online learning algorithm which utilizes online passive-aggressive learning (PA) and total-error-rate  minimization (TER) for binary classification.
	The PA learning establishes not only large margin training but also the capacity to handle non-separable data.
	The TER learning on the other hand minimizes an approximated classification error based objective function.
	We propose an online PATER algorithm which combines those useful properties.
	In addition, we also present a weighted PATER algorithm to improve the ability to cope with data imbalance problems.
	Experimental results demonstrate that the proposed PATER algorithms achieves better performances in terms of efficiency and effectiveness than the existing state-of-the-art online learning algorithms in real-world data sets.
\end{abstract}

\section{Introduction}
Online learning has been widely studied and efficiently applied to sequentially arriving data problems that cannot be solved by batch learning \cite{cesa2006prediction, shalev2011online, hoi2018online}. In online binary classification, Perceptron \cite{rosenblatt1958perceptron} is one of the most popular algorithms based on the first-order information. The  Perceptron adopts the following step loss function to minimize misclassification:
\begin{equation} \label{eq.l_sign}
\ell_{step}(\pmb{w}) =\left\{\begin{array}{ c c }
0 & \mathbf{\pmb{w} \cdot x}^{}_{t} \geq 0\\
1 & \mathbf{\pmb{w} \cdot x}^{}_{t} < 0
\end{array} \right. ,
\end{equation}
where $\pmb{w} \in \mathbb{R}^{d}$ is a parameter vector, and $\mathbf{x}_{t} \in \mathbb{R}^{d}$ is a sample vector at time $t$.
In view of margin based learning, online passive-aggressive learning  (PA)  \cite{crammer2006online} has been successfully developed as a notable online learning method. The PA updates their learning parameters with a large margin obtained by the following hinge loss function:
\begin{equation} \label{eq.hinge_loss}
\ell_{hinge}(\pmb{w}) =\left\{\begin{array}{ c c }
0  											 		& y_{t}\left(\mathbf{\pmb{w}\cdot x}_{t}\right) \geq 1\\
1-y_{t}\left(\mathbf{\pmb{w}\cdot x}_{t}\right) 	& y_{t}\left(\mathbf{\pmb{w}\cdot x}_{t}\right) < 1
\end{array} \right. ,
\end{equation}
where $\displaystyle y_{t} \in \{-1,1\}$ is a target label.
The PA also accommodates non-separable data problems by not assuming data separability.
In the use of the second-order information, 
the second-order perceptron \cite{cesa2005second}, the confidence weighted learning \cite{dredze2008confidence}, the adaptive regularization of weights learning \cite{crammer2009adaptive} have received considerable attention.
Although the second-order information ensures the faster optimization convergence, the use of the first-order information is still reasonably attractive due to its simplicity.

Different from the margin based learning, Total-Error-Rate minimization (TER) based learning \cite{toh2008deterministic} has been introduced  with a classification error approximation for binary classification (see Section \ref{sec.TER}). The TER takes two different step loss functions for false negative and false positive involved in the confusion matrix of binary classification. In order to achieve a deterministic global solution and avoid local solutions, the TER objective function adopts a quadratic approximation to the step function for the desired convexity.  
However, the above TER based solution has constructed a batch learning solution that has  poor scalability for large-scale and real-time applications.

In this paper, we introduce a new online learning algorithm
which can put in use the properties of PA and TER and overcome their individual shortcomings by combining them in which we call PATER for online binary classification. 
In addition, motivated by a weighted accuracy scheme \cite{brodersen2010balanced}, we also propose a weighted PATER (wPATER) method for data imbalance problems.
Our empirical results demonstrate that the proposed methods show promising performances on 31 real-world data sets in terms of efficiency and effectiveness.

%Least-squares (LS) minimization has been rigorously investigated based on a quadratic loss function (e.g., the sum of the squared difference) between the true and the predicted labels for not only regression but also classification. 
%Although LS based methods are successfully applied to classification problems, the design of the LS objective function is still fitted into regression problems.

\section{Preliminaries}
\subsection{Online Passive-Aggressive Learning (PA)}
We introduce online Passive-Aggressive learning (PA) which is one of the popular online learning methods of linear classifiers for binary classification \cite{crammer2006online}. The optimization problem of the PA  is as follows:
\begin{equation} \label{eq.PA_objective}
{{\pmb{w}}_{t + 1}} = \arg \min \frac{1}{2}{\left\| {{\pmb{w}} - {{\pmb{w}}_t}} \right\|^2}   \text{~~~} s.t.  \text{~~~} \ell_{hinge}(\pmb{w}) = 0,
\end{equation}
%where $\pmb{w}$ is a parameter vector, and $\ell_{hinge}(\pmb{w})$ is the hinge loss function as:
%\begin{equation} \label{eq.hinge_loss}
%\ell_{hinge}(\pmb{w}) =\left\{\begin{array}{ c c }
%0  											 		& y_{t}\left(\mathbf{\pmb{w}\cdot x}_{t}\right) \geq 1\\
%1-y_{t}\left(\mathbf{\pmb{w}\cdot x}_{t}\right) 	& y_{t}\left(\mathbf{\pmb{w}\cdot x}_{t}\right) < 1
%\end{array} \right. .
%\end{equation}
%$\displaystyle \mathbf{x}_{t} \in \mathbb{R}^{d}$ is a sample vector, and $\displaystyle y_{t} \in \{-1,1\}$ is a target label at time $t$. 
%$d$ is a feature dimension. 
By solving this optimization problem, the PA solution is given by:
\begin{equation}
{{\pmb{w}}_{t + 1}} = {{\pmb{w}}_t} + \tau_t y_t \mathbf{x}_t   ,
\end{equation}
where $\tau_t = \frac{\ell_{hinge}(\pmb{w})}{\left\| \mathbf{x}_t \right\|^2}  $ is a Lagrange multiplier under the Karush-Khun-Tucker (KKT) condition \cite{boyd2004convex}. The online PA solution is passively updated when the loss is close zero,  whereas it is aggressively updated when it suffered from a significant loss.

\subsection{Total-Error-Rate (TER) Minimization}
\label{sec.TER}

%The Least-Squares (LS) minimization is the most common method for both regression and classification. The objective function of the LS minimization   is based on the sum of squared errors  distance function more relevant to the regression problems as follows:
%\begin{equation} \label{eq.LSE_J}
%\begin{aligned}
%\text{LS: } J &= \frac{1}{2 }\sum\limits_{i = 1}^n {{{\left( {{y_i} - {\pmb{w }}^T{{\bf{x}}_i}} \right)}^2}},\end{aligned}
%\end{equation}
%which provides a deterministic closed-form solution as
%$\pmb{w} = (\mathbf{X}^T\mathbf{X} +
%b{\bf{I}})^{-1}\mathbf{X}^T\mathbf{y}$, 
%where $\mathbf{X} \in \mathbb{R}^{n\times d}$ is a data matrix, $\mathbf{y} \in  \{-1, 1\}$ is the target label vector, $n$ indicates the number of data samples, and $d$ indicates a sample feature dimension. 
%$b$ is a regularization factor, and ${\bf I}$ is an identity matrix with similar dimension as that of ${\bf X}^T {\bf X}$.

In \cite{toh2008deterministic}, a Total-Error-Rate (TER) has been utilized in an optimization problem for binary classification. The TER is defined as the sum of the False Positive Rate (FPR) and the False Negative Rate (FNR):
\begin{equation} \label{eq.l_step_TER}
\begin{aligned}
\ell _{step}^{TER}(\pmb{w}) &= FPR + FNR \\
&=\frac{1}{n^{-}}\sum ^{n^{-}}_{i=1} \ell _{step}^{FP}(\pmb{w}) +\frac{1}{n^{+}}\sum ^{n^{+}}_{j=1} \ell _{step}^{FN}(\pmb{w}) ,
\end{aligned}
\end{equation}
where
\begin{equation} \label{eq.l_step_FP_FN}
%\small
\begin{array}{c}
\ell_{step}^{FP}(\pmb{w}) =\left\{\begin{array}{ c c }
0 & -\mathbf{\pmb{w} \cdot x}^{-}_{t} \geq 0\\
1 & -\mathbf{\pmb{w} \cdot x}^{-}_{t} < 0
\end{array} \right. ,\\ \\
\ell_{step}^{FN}(\pmb{w}) =\left\{\begin{array}{ c c }
0 & \phantom{-}\mathbf{\pmb{w} \cdot x}^{+}_{t} \geq 0\\
1 & \phantom{-}\mathbf{\pmb{w} \cdot x}^{+}_{t} < 0
\end{array} \right. .
\end{array}
\end{equation}
$n^-$ and $n^+$ is the numbers of negative and positive samples respectively $(n = n^- + n^+)$.
The two superscripts, $-$ and $+$, are for  indication of each class.
%the sum of False Positive (FP) rate, $\text{FP} \over n^- $ and False Negative (FN) rate, $\text{FN} \over n^+ $, where $n^-$ and $n^+$ is the numbers of negative and positive samples respectively $(n = n^- + n^+)$.
The TER objective function has been presented based on the quadratic approximation to the two step loss functions of FP and FN as follows: 
\begin{equation} \label{eq.TER_J}
\footnotesize
\begin{aligned} {{\pmb{w}}} &= \arg \min \frac{1}{n^{-}}\sum ^{n^{-}}_{i=1} \ell _{quad}^{FP}(\pmb{w}) +\frac{1}{n^{+}}\sum ^{n^{+}}_{j=1} \ell _{quad}^{FN}(\pmb{w}) ,
\\
&= \arg \min \frac{1}{{2n_{}^ - }}\sum\limits_{i = 1}^{n_{}^ - } {{{\left( {y_i^ -  - ({\pmb{w }}\cdot{\bf{x}}_i^ -) } \right)}^2}}  + \frac{1}{{2n_{}^ + }}\sum\limits_{j = 1}^{n_{}^ + } {{{\left( {y_j^ +  - ({\pmb{w }}\cdot{\bf{x}}_j^ +) } \right)}^2}}, 
\end{aligned}
\end{equation}
where 
\begin{equation} \label{eq.l_quad_FP_FN}
%\small
\begin{array}{c}
\ell_{quad}^{FP}(\pmb{w}) = {{{\left( {y^ -  - ({\pmb{w }}\cdot{\bf{x}}^ -) } \right)}^2}} , \\ \\
\ell_{quad}^{FN}(\pmb{w}) = {{{\left( {y^ +  - ({\pmb{w }}\cdot{\bf{x}}^ +) } \right)}^2}} ,
\end{array}
\end{equation}
which yields a closed-form solution related to weighted least-squares as 
\begin{equation} \label{eq.TER_sol}
\pmb{w} = (\mathbf{X}^T\mathbf{WX} )^{-1}\mathbf{X}^T\mathbf{Wy},
\end{equation}
where $\mathbf{X} =\left[
\mathbf{X}^{-},  \mathbf{X}^{+}
\right]^{T}$ includes two sample matrices for negative and positive classes.
%$b$ is a regularization factor, and ${\bf I}$ is an identity matrix with similar dimension as that of ${\bf X}^T {\bf WX}$.
$\mathbf{W} =
diag\begin{pmatrix}\begin{bmatrix}\frac{1}{n^-}\,,\,\dotsc\,,\,\frac{1}{n^-}\,,\,\frac{1}{n^+}\,,\,\dotsc\,,\,\frac{1}{n^+}
\end{bmatrix}\end{pmatrix}$ is a weight matrix that includes two different weights for the negative and positive classes.  
$\mathbf{y} = \begin{bmatrix} -1,\,\dotsc,\,-1,\,1,\,\dotsc,\,1\end{bmatrix}$ is the target vector.
The batch mode TER solution handles binary classification problems by approximately counting the misclassified samples using the two different  weights, $1 \over n^-$  and $1 \over n^+$. 

\section{Online Passive-Aggressive Total-Error-Rate Minimization (PATER)}
\label{sec.PROPOSED1}

In this section, based on the online PA learning, we present an online TER minimization algorithm  (PATER).
We start by addressing the two step loss functions $\ell_{step}^{FP}(\pmb{w})$ and $\ell_{step}^{FN}(\pmb{w})$ of TER.
Similar to the hinge loss function in \eqref{eq.hinge_loss}, these $\ell_{step}^{FP}(\pmb{w})$ and $\ell_{step}^{FN}(\pmb{w})$ can be rewritten as:
\begin{equation} \label{eq.l_hinge_FP_FN}
%\small
\begin{array}{c}
\ell_{hinge}^{FP}(\pmb{w}) =\left\{\begin{array}{ c c }
0                                  	& -\mathbf{\pmb{w} \cdot x}^{-}_{t} \geq 1\\
1+\mathbf{\pmb{w} \cdot x}^{-}_{t} 	& -\mathbf{\pmb{w} \cdot x}^{-}_{t}    < 1
\end{array} \right. ,\\ \\
\ell_{hinge}^{FN}(\pmb{w}) =\left\{\begin{array}{ c c }
0  									& \phantom{-}\mathbf{\pmb{w} \cdot x}^{+}_{t} \geq 1\\
1-\mathbf{\pmb{w} \cdot x}^{+}_{t} 	& \phantom{-}\mathbf{\pmb{w} \cdot x}^{+}_{t}    < 1
\end{array} \right. .
\end{array}
\end{equation}
Based on $\ell_{hinge}^{FP}(\pmb{w})$ and $\ell_{hinge}^{FN}(\pmb{w})$ in \eqref{eq.l_hinge_FP_FN}, a new TER loss function is given by:
\begin{equation} \label{eq.l_hinge_TER}
\ell _{hinge}^{TER}(\pmb{w}) =\frac{1}{n^{-}}\sum ^{n^{-}}_{i=1} \ell _{hinge}^{FP}(\pmb{w}) +\frac{1}{n^{+}}\sum ^{n^{+}}_{j=1} \ell _{hinge}^{FN}(\pmb{w})
\end{equation}

Similar to the PA objective function in \eqref{eq.PA_objective}, a new TER objective function using \eqref{eq.l_hinge_TER} can be written as:
\begin{equation} \label{eq.TER_objective}
{{\pmb{w}}_{t + 1}} = \arg \min \frac{1}{2}{\left\| {{\pmb{w}} - {{\pmb{w}}_t}} \right\|^2}   \text{~~~} s.t.  \text{~~~} \ell _{hinge}^{TER}(\pmb{w})  = 0.
\end{equation}
The Lagrangian of optimization problem for TER minimization can then be defined as follows:
\begin{equation}  \label{eq.Lagrangian_TER}
%\small
\begin{aligned}
\mathcal{L}( \pmb{w},\tau ) &= \frac{1}{2}\Vert \pmb{w}-\pmb{w}_{t}\Vert ^{2}\\
&\phantom{=} +\tau _{t}\left(\frac{1}{n^{-}_{t}}\sum ^{n^{-}_{t}}_{i=1} 1+\mathbf{\pmb{w}\cdot x}^{-}_{i} +\frac{1}{n^{+}_{t}}\sum ^{n^{+}_{t}}_{j=1} 1-\mathbf{\pmb{w}\cdot x}^{+}_{j}\right) .
\end{aligned}
\end{equation}
Taking the derivative of $\mathcal{L}( \pmb{w},\tau )$ with respect to $\pmb{w}$ and setting it to zero as:
\begin{equation}
%\small
\nabla _{\pmb{w}}\mathcal{L}( \pmb{w},\tau ) =\pmb{w}-\pmb{w}_{t} +\tau _{t}\left(\frac{1}{n^{-}_{t}}\sum ^{n^{-}_{t}}_{i=1}\mathbf{x}^{-}_{i} -\frac{1}{n^{+}_{t}}\sum ^{n^{+}_{t}}_{j=1}\mathbf{x}^{+}_{j}\right) =0 ,
\end{equation}
the PATER solution $\pmb{w}$ is given by:
\begin{equation} \label{eq.PATER_solution}
\pmb{w}=\pmb{w}_{t} -\tau _{t}\left(\frac{1}{n^{-}_{t}}\sum ^{n^{-}_{t}}_{i=1}\mathbf{x}^{-}_{i} -\frac{1}{n^{+}_{t}}\sum ^{n^{+}_{t}}_{j=1}\mathbf{x}^{+}_{j}\right) .
\end{equation}
In order to transform the two summation terms into vectorized recursive forms in \eqref{eq.PATER_solution}, 
the PATER solution $\pmb{w}$ can be rewritten as:
\begin{equation}\label{eq.PATER_solution2}
%\small
\begin{aligned}
\pmb{w} &= \pmb{w}_{t} -\tau _{t}\left(\left(\frac{n^{-}_{t-1}}{n^{-}_{t}}\sum ^{n^{-}_{t-1}}_{i=1}\frac{1}{n^{-}_{t-1}}\mathbf{x}^{-}_{i} +\lambda ^{-}_{t}\frac{1}{n^{-}_{t}}\mathbf{x}^{}_{t}\right)  \right.\\
&\phantom{=} \left. -\left(\frac{n^{+}_{t-1}}{n^{+}_{t}}\sum ^{n^{+}_{t-1}}_{j=1}\frac{1}{n^{+}_{t-1}}\mathbf{x}^{+}_{j} +\lambda ^{+}_{t}\frac{1}{n^{+}_{t}}\mathbf{x}^{}_{t}\right) \right)\\
&= \pmb{w}_{t} -\tau _{t}\left(\left(\frac{n^{-}_{t-1}}{n^{-}_{t}}\mathbf{z}^{-}_{t-1} +\frac{\lambda ^{-}_{t}}{n^{-}_{t}}\mathbf{x}^{}_{t}\right) -\left(\frac{n^{+}_{t-1}}{n^{+}_{t}}\mathbf{z}^{+}_{t-1} +\frac{\lambda ^{+}_{t}}{n^{+}_{t}}\mathbf{x}^{}_{t}\right)\right)\\
&= \pmb{w}_{t} -\tau _{t}\left(\mathbf{z}^{-}_{t} -\mathbf{z}^{+}_{t}\right)\\
&= \pmb{w}_{t} +\tau _{t}\mathbf{z}_{t} ,
\end{aligned}
\end{equation}
where  $ \lambda ^{-}_{t} =\frac{1-y_{t}}{2}$ and $ \lambda ^{+}_{t} =\frac{1+y_{t}}{2}$ are  indicators to select one class between the two classes at time $t$. 
$ \mathbf{z}_{t} =\mathbf{z}^{+}_{t} - \mathbf{z}^{-}_{t}$, 
$ \mathbf{z}^{-}_{t} =\frac{n^{-}_{t-1}}{n^{-}_{t}}\mathbf{z}^{-}_{t-1} +\frac{\lambda ^{-}_{t}}{n^{-}_{t}}\mathbf{x}^{}_{t}$, and
$ \mathbf{z}^{+}_{t} =\frac{n^{+}_{t-1}}{n^{+}_{t}}\mathbf{z}^{+}_{t-1} +\frac{\lambda ^{+}_{t}}{n^{+}_{t}}\mathbf{x}^{}_{t}$.

Substituting the parameter vector $\pmb{w}$ with \eqref{eq.PATER_solution2} in \eqref{eq.Lagrangian_TER}, we get
\begin{equation} \label{eq.Lagrangian2_TER}
\begin{aligned}
\mathcal{L}( \tau ) & =-\frac{1}{2} \tau ^{2}\left\Vert \mathbf{z}_{t}\right\Vert ^{2}\\
&\phantom{=} +\tau \left(\frac{1}{n^{-}_{t}}\sum ^{n^{-}_{t}}_{i=1} 1+\pmb{w}_{t}\mathbf{\cdot x}^{-}_{i} +\frac{1}{n^{+}_{t}}\sum ^{n^{+}_{t}}_{j=1} 1-\pmb{w}_{t}\mathbf{\cdot x}^{+}_{j}\right) .
\end{aligned}
\end{equation}
In \eqref{eq.Lagrangian2_TER}, the linearly increasing computational complexity $ \mathcal{O}( dn_{t})$ of the summation terms is observed and caused by the dot product of $\pmb{w}_{t}\mathbf{\cdot x}$ for each sample. 
To avoid this computation, two different versions, namely PATER-I and PATER-II, of \eqref{eq.Lagrangian2_TER} are given by:
%\begin{equation} \label{eq.PATER1}
%\small
%\begin{aligned}
%\text{PATER-I: }\mathcal{L}( \tau ) & =-\frac{1}{2} \tau ^{2}\left\Vert \mathbf{z}_{t}\right\Vert ^{2}\\
%&\phantom{=} +\tau \left( \lambda ^{-}_{t}\frac{1}{n^{-}_{t}}\left( 1+w_{t}\mathbf{\cdot x}^{-}_{t}\right) +\lambda ^{+}_{t}\frac{1}{n^{+}_{t}}\left( 1-w_{t}\mathbf{\cdot x}^{+}_{t}\right)\right) ,
%\end{aligned}
%\end{equation}
\begin{equation} \label{eq.PATER1}
%\small
\begin{aligned}
\text{PATER-I: }\mathcal{L}( \tau ) & =-\frac{1}{2} \tau ^{2}\left\Vert \mathbf{z}_{t}\right\Vert ^{2} +\tau \left( \frac{\lambda ^{-}_{t}}{n^{-}_{t}}\left( 1+\pmb{w}_{t}\mathbf{\cdot x}^{}_{t}\right)\right. \\
&\phantom{=} \left. +\frac{\lambda ^{+}_{t}}{n^{+}_{t}}\left( 1-\pmb{w}_{t}\mathbf{\cdot x}^{}_{t}\right)\right) ,
\end{aligned}
\end{equation}
\begin{equation} \label{eq.PATER2}
%\small
\begin{aligned}
\text{PATER-II: }\mathcal{L}( \tau ) & =-\frac{1}{2} \tau ^{2}\left\Vert \mathbf{z}_{t}\right\Vert ^{2} +\tau \left(\frac{n^{-}_{t-1}}{n^{-}_{t}} k^{-}_{t-1} +\frac{\lambda ^{-}_{t}}{n^{-}_{t}}\left( 1+\pmb{w}_{t}\mathbf{\cdot x}^{}_{t}\right)\right. \\
&\phantom{=} \left. +\frac{n^{+}_{t-1}}{n^{+}_{t}} k^{+}_{t-1} +\frac{\lambda ^{+}_{t}}{n^{+}_{t}}\left( 1-\pmb{w}_{t}\mathbf{\cdot x}^{}_{t}\right)\right)\\
&= -\frac{1}{2} \tau ^{2}\left\Vert \mathbf{z}_{t}\right\Vert ^{2} +\tau \left( k^{-}_{t} +k^{+}_{t}\right) ,
\end{aligned}
\end{equation}
where $ k^{-}_{0} =k^{+}_{0} =0$,
$ k^{-}_{t} =\frac{n^{-}_{t-1}}{n^{-}_{t}} k^{-}_{t-1} +\frac{\lambda ^{-}_{t}}{n^{-}_{t}}\left( 1+\pmb{w}_{t}\mathbf{\cdot x}^{}_{t}\right)$, and
$ k^{+}_{t} =\frac{n^{+}_{t-1}}{n^{+}_{t}} k^{+}_{t-1} +\frac{\lambda ^{+}_{t}}{n^{+}_{t}}\left( 1-\pmb{w}_{t}\mathbf{\cdot x}^{}_{t}\right)$.

Taking the derivative of $\mathcal{L}( \tau )$ in \eqref{eq.PATER1} and \eqref{eq.PATER2} with respect to $\tau$ and setting it zero as:
\begin{equation}  
%\small
\begin{aligned}
\text{PATER-I: }\frac{\partial \mathcal{L}( \tau )}{\partial \tau } & =-\tau \left\Vert \mathbf{z}_{t}\right\Vert ^{2} +\frac{\lambda ^{-}_{t}}{n^{-}_{t}}\left( 1+\pmb{w}_{t}\mathbf{\cdot x}^{}_{t}\right)\\
&\phantom{=} +\frac{\lambda ^{+}_{t}}{n^{+}_{t}}\left( 1-\pmb{w}_{t}\mathbf{\cdot x}^{}_{t}\right)=0 ,
\end{aligned}
\end{equation}
\begin{equation} 
%\small
\text{PATER-II: }\frac{\partial \mathcal{L}( \tau )}{\partial \tau } =-\tau \left\Vert \mathbf{z}_{t}\right\Vert ^{2} +k^{-}_{t} +k^{+}_{t}=0 ,
\end{equation}
we get:
\begin{equation} \label{eq.PATER1_tau}
%\small
\text{PATER-I: } \tau _{t} =\frac{\frac{\lambda ^{-}_{t}}{n^{-}_{t}}\left( 1+\pmb{w}_{t}\mathbf{\cdot x}^{}_{t}\right) +\frac{\lambda ^{+}_{t}}{n^{+}_{t}}\left( 1-\pmb{w}_{t}\mathbf{\cdot x}^{}_{t}\right)}{\left\Vert \mathbf{z}_{t}\right\Vert ^{2}} ,
\end{equation}
\begin{equation} \label{eq.PATER2_tau}
%\small
\text{PATER-II: } \tau _{t} =\frac{k^{-}_{t} +k^{+}_{t}}{\left\Vert \mathbf{z}_{t}\right\Vert ^{2}} .
\end{equation}

\begin{algorithm}[t]
	\caption{Passive-Aggressive TER Minimization}
	\label{alg.PATER1}
	\begin{algorithmic}
		\STATE {\bfseries Input:} a sample $\mathbf{x}_t \in \mathbb{R}^{d}$, \\
		\hskip3.1em a label $y_t \in \{-1,+1\}$
		\STATE {\bfseries Initialize:} $\pmb{w}_0= \mathbf{z}_0^-=\mathbf{z}_0^+=\mathbf{0}$, \\
		\hskip4.5em $n_0^-=n_0^+=k_t^-=k_t^+=0$
		%$\pmb{w}_0=\mathbf{0}$, $\mathbf{z}_0^-=\mathbf{z}_0^+=\mathbf{0}$, $n_0^-=n_0^+=0$

		\FOR{$t=1,\dots$ }
		\STATE Set $ \lambda ^{-}_{t} =\frac{1-y_{t}}{2}$ and $ \lambda ^{+}_{t} =\frac{1+y_{t}}{2}$
		\STATE Update $n_t^-=n_{t-1}^- + \lambda ^{-}_{t}$,  \\ 
		\hskip3.1em $n_t^+=n_{t-1}^+ + \lambda ^{+}_{t}$
		\STATE Update $ \mathbf{z}^{-}_{t} =\frac{n^{-}_{t-1}}{n^{-}_{t}}\mathbf{z}^{-}_{t-1} +\frac{\lambda ^{-}_{t}}{n^{-}_{t}}\mathbf{x}^{}_{t}$, \\
		\hskip3.1em $ \mathbf{z}^{+}_{t} =\frac{n^{+}_{t-1}}{n^{+}_{t}}\mathbf{z}^{+}_{t-1} +\frac{\lambda ^{+}_{t}}{n^{+}_{t}}\mathbf{x}^{}_{t}$
		\STATE Update $ k^{-}_{t} =\frac{n^{-}_{t-1}}{n^{-}_{t}} k^{-}_{t-1} +\frac{\lambda ^{-}_{t}}{n^{-}_{t}}\left( 1+\pmb{w}_{t-1}\mathbf{\cdot x}^{}_{t}\right)$, \\
		\hskip3.1em $ k^{+}_{t} =\frac{n^{+}_{t-1}}{n^{+}_{t}} k^{+}_{t-1} +\frac{\lambda ^{+}_{t}}{n^{+}_{t}}\left( 1-\pmb{w}_{t-1}\mathbf{\cdot x}^{}_{t}\right)$
		\STATE Set $\mathbf{z}_{t} = \mathbf{z}^{+}_{t} - \mathbf{z}^{-}_{t}$
		\STATE Set $\tau _{t}$ as:  \\
		\hskip2.2em PATER-I:  $\tau _{t} = \frac{\frac{\lambda ^{-}_{t}}{n^{-}_{t}}\left( 1+\pmb{w}_{t-1}\mathbf{\cdot x}^{}_{t}\right) +\frac{\lambda ^{+}_{t}}{n^{+}_{t}}\left( 1-\pmb{w}_{t-1}\mathbf{\cdot x}^{}_{t}\right)}{\left\Vert \mathbf{z}_{t}\right\Vert ^{2}}$ \\  
		\hskip2.2em PATER-II: $\tau _{t} =\frac{k^{-}_{t} +k^{+}_{t}}{\left\Vert \mathbf{z}_{t}\right\Vert ^{2}}$
		
		%		\IF{$x_i > x_{i+1}$}
		%		\STATE Swap $x_i$ and $x_{i+1}$
		%		\STATE $noChange = false$
		%		\ENDIF
		\STATE Update $\pmb{w}_{t} = \pmb{w}_{t-1} +\tau _{t}\mathbf{z}_{t}$
		\ENDFOR
		
	\end{algorithmic}
\end{algorithm}

\subsection{Summary}
The pseudo code of the proposed PATER algorithm is given in Algorithm \ref{alg.PATER1}.
The main difference from the PA learning is that the PATER solution is given by the newly derived objective function \eqref{eq.TER_objective} utilizing the PA and the TER learning together.
The main difference between PATER-I and PATER-II  is that PATER-I includes an instantaneous loss while PATER-II uses an approximately accumulated loss in $\tau _{t}$.

\section{Weighted PATER Minimization (wPATER)}
\label{sec.PROPOSED2}
In this section, we present a weighted PATER algorithm (wPATER) inspired by the concept of a balanced accuracy for performance evaluation in \cite{brodersen2010balanced}. 
In order to address  imbalanced data sets, a balanced accuracy is defined as 
$\displaystyle bAcc=\frac{1}{2}\left(\frac{TN}{n^{-}} +\frac{TP}{n^{+}}\right)$, 
where TN is the True Negative, and TP is the True Positive. 
In view of a generalized design, a weighted accuracy can be written as: 
\begin{equation} \label{eq.def_wTER}
wAcc= \alpha^- \frac{TN}{n^{-}} + \alpha^+ \frac{TP}{n^{+}},
\end{equation}
where $bAcc$ is given by $\alpha^- = \alpha^+ = 0.5$.
In a similar manner to this weighted accuracy, a weighted TER (wTER)  can be defined as 
\begin{equation} \label{eq.def_wTER}
wTER = \alpha^- \frac{FP}{n^{-}} + \alpha^+  \frac{FN}{n^{+}}.
\end{equation}
Based on the wTER definition, similar to \eqref{eq.l_hinge_TER}, 
a new wTER loss function is given by:
\begin{equation} \label{eq.l_hinge_wTER}
\ell _{hinge}^{wTER}(\pmb{w}) =\frac{\alpha^- }{n^{-}}\sum ^{n^{-}}_{i=1} \ell _{hinge}^{FP}(\pmb{w}) +\frac{\alpha^+ }{n^{+}}\sum ^{n^{+}}_{j=1} \ell _{hinge}^{FN}(\pmb{w}) .
\end{equation}
Accordingly, the wTER objective function can be written as:
\begin{equation} \label{eq.wTER_objective}
{{\pmb{w}}_{t + 1}} = \arg \min \frac{1}{2}{\left\| {{\pmb{w}} - {{\pmb{w}}_t}} \right\|^2}   \text{~~~} s.t.  \text{~~~} \ell_{hinge}^{wTER}(\pmb{w}) = 0.
\end{equation}
Similar to \eqref{eq.Lagrangian_TER}, the Lagrangian of the optimization problem can then be defined as:
\begin{equation}  \label{eq.Lagrangian_wTER}
%\small
\begin{aligned}
\mathcal{L}( \pmb{w},\tau ) &= \frac{1}{2}\Vert \pmb{w}-\pmb{w}_{t}\Vert ^{2}\\
&\phantom{=} + \tau _{t}\left(\frac{\alpha^-}{n^{-}_{t}}\sum ^{n^{-}_{t}}_{i=1} 1+\mathbf{\pmb{w}\cdot x}^{-}_{i} +\frac{\alpha^+}{n^{+}_{t}}\sum ^{n^{+}_{t}}_{j=1} 1-\mathbf{\pmb{w}\cdot x}^{+}_{j}\right) .
\end{aligned}
\end{equation}
Similarly, the wPATER solution $\pmb{w}$ is given by:
\begin{equation} \label{eq.wPATER_solution}
\pmb{w}=\pmb{w}_{t} + \tau _{t}\mathbf{z}_{t},
\end{equation}
where $\mathbf{z}_{t} = \alpha^+ \mathbf{z}^{+}_{t} - \alpha^- \mathbf{z}^{-}_{t}$.
Next, similar to \eqref{eq.Lagrangian2_TER}, we get:
\begin{equation} \label{eq.Lagrangian2_wTER}
\begin{aligned}
\mathcal{L}( \tau ) & =-\frac{1}{2} \tau^2\left\Vert \mathbf{z}_{t}\right\Vert ^{2}\\
&\phantom{=} + \tau \left(\frac{\alpha^-}{n^{-}_{t}}\sum ^{n^{-}_{t}}_{i=1} 1+\pmb{w}_{t}\mathbf{\cdot x}^{-}_{i} +\frac{\alpha^+}{n^{+}_{t}}\sum ^{n^{+}_{t}}_{j=1} 1-\pmb{w}_{t}\mathbf{\cdot x}^{+}_{j}\right) .
\end{aligned}
\end{equation}

\begin{algorithm}[t]
	\caption{Weighted PATER Minimization}
	\label{alg.wPATER1}
	\begin{algorithmic}
		\STATE {\bfseries Input:} a sample $\mathbf{x}_t \in \mathbb{R}^{d}$, \\
		\hskip3.1em a label $y_t \in \{-1,+1\}$
		\STATE {\bfseries Initialize:} $\pmb{w}_0= \mathbf{z}_0^-=\mathbf{z}_0^+=\mathbf{0}$, \\
		\hskip4.5em $n_0^-=n_0^+=k_t^-=k_t^+=0$, \\
		\hskip4.5em $\alpha^- > 0, \alpha^+ > 0$
		%$\pmb{w}_0=\mathbf{0}$, $\mathbf{z}_0^-=\mathbf{z}_0^+=\mathbf{0}$, $n_0^-=n_0^+=0$

		\FOR{$t=1,\dots$ }
		\STATE Set $ \lambda ^{-}_{t} =\frac{1-y_{t}}{2}$ and $ \lambda ^{+}_{t} =\frac{1+y_{t}}{2}$
		\STATE Update $n_t^-=n_{t-1}^- + \lambda ^{-}_{t}$, \\
		\hskip3.1em  $n_t^+=n_{t-1}^+ + \lambda ^{+}_{t}$
		\STATE Update $ \mathbf{z}^{-}_{t} =\frac{n^{-}_{t-1}}{n^{-}_{t}}\mathbf{z}^{-}_{t-1} +\frac{\lambda ^{-}_{t}}{n^{-}_{t}}\mathbf{x}^{}_{t}$, \\
		\hskip3.1em $ \mathbf{z}^{+}_{t} =\frac{n^{+}_{t-1}}{n^{+}_{t}}\mathbf{z}^{+}_{t-1} +\frac{\lambda ^{+}_{t}}{n^{+}_{t}}\mathbf{x}^{}_{t}$
		\STATE Update $ k^{-}_{t} =\frac{n^{-}_{t-1}}{n^{-}_{t}} k^{-}_{t-1} +\frac{\lambda ^{-}_{t}}{n^{-}_{t}}\left( 1+\pmb{w}_{t-1}\mathbf{\cdot x}^{}_{t}\right)$, \\
		\hskip3.1em $ k^{+}_{t} =\frac{n^{+}_{t-1}}{n^{+}_{t}} k^{+}_{t-1} +\frac{\lambda ^{+}_{t}}{n^{+}_{t}}\left( 1-\pmb{w}_{t-1}\mathbf{\cdot x}^{}_{t}\right)$
		\STATE Set $\mathbf{z}_{t} = \alpha^+ \mathbf{z}^{+}_{t} - \alpha^- \mathbf{z}^{-}_{t}$
		\STATE Set $\tau _{t}$ as:  \\
		\hskip2.2em wPATER-I:  $\tau _{t} = \frac{\frac{\alpha^- \lambda ^{-}_{t}}{n^{-}_{t}}\left( 1+\pmb{w}_{t-1}\mathbf{\cdot x}^{}_{t}\right) +\frac{\alpha^+ \lambda ^{+}_{t}}{n^{+}_{t}}\left( 1-\pmb{w}_{t-1}\mathbf{\cdot x}^{}_{t}\right)}{\left\Vert \mathbf{z}_{t}\right\Vert ^{2}}$ \\  
		\hskip2.2em wPATER-II: $\tau _{t} =\frac{\alpha^- k^{-}_{t} +\alpha^+ k^{+}_{t}}{\left\Vert \mathbf{z}_{t}\right\Vert ^{2}}$
		
		%		\IF{$x_i > x_{i+1}$}
		%		\STATE Swap $x_i$ and $x_{i+1}$
		%		\STATE $noChange = false$
		%		\ENDIF
		\STATE Update $\pmb{w}_{t} = \pmb{w}_{t-1} +\tau _{t}\mathbf{z}_{t}$
		\ENDFOR
		
	\end{algorithmic}
\end{algorithm}

\begin{table*}[t!]
	\caption{Summary of the 31 real-world data sets for binary classification.} \label{tbl.dataSummary}
	\vskip 0.15in
	\begin{center}
		\begin{small}
			\begin{sc}
				\fontsize{6}{8}\selectfont
				\centering 
				\begin{tabular}{c@{\hskip0.5pt}llllllc@{\hskip3pt}llllll}
					\hline
					%\multirow{27}{*}{\rotatebox[origin=c]{90}{Small and Medium }}
					& No. {\rule{0pt}{3ex}} & Data sets          & \#cases(original)    & \#feat & Ratio & \#miss & No. {\rule{0pt}{3ex}} & Data sets          & \#cases(original)    & \#feat & Ratio  & \#miss   \\ \hline
					& 1 
					{\rule{0pt}{3ex}}  		   
					& Monks-3     	   & 122(423)   & 6      & 0.98       & No      
					& 17  & Blood-transfusion  & 748        & 4      & 0.31       & No 			\\
					
					& 2   & Monks-1            & 124(423)   & 6      & 1.01       & No       
					& 18  & Pima-diabetes      & 768        & 8      & 0.54       & No				\\
					
					& 3   & Monks-2            & 169(423)   & 6      & 0.62       & No       
					& 19  & Mammographic       & 830(961)   & 5      & 0.95       & 131			\\
					
					& 4   & Wpbc               & 194(198)   & 33     & 0.31       & 4        
					& 20  & Tic-tac-toe        & 958        & 9      & 0.53       & No				\\
					
					& 5   & Parkinsons         & 195        & 22     & 3.15       & No       
					& 21  & Statlog-german     & 1,000      & 24     & 2.34       & No 			\\
					
					& 6   & Sonar              & 208        & 60     & 1.16       & No       
					& 22  & Ozone-eight        & 1,847(2,534) & 72   & 0.07       & 687 			\\
					
					& 7   & SPECTF-heart       & 267        & 44     & 3.89       & No       
					& 23  & Ozone-one          & 1,848(2,536) & 72   & 0.03       & 688 			\\
					
					& 8   & StatLog-heart      & 270        & 13     & 0.80       & No       
					& 24  & 20News-talk        & 1,848 	    & 3      & 1.04       & No 			\\
					
					%		9   & Shuttle-l-control  & 279(15)    & 6      & 2       & No       \\
					& 9   & BUPA-liver         & 345        & 6      & 1.39       & No       
					& 25  & 20News-comp        & 1,937      & 3      & 0.98       & No 			\\
					
					& 10  & Ionosphere         & 351        & 34     & 0.56       & No       
					& 26  & 20News-sci         & 1,971      & 3      & 1.01       & No 			\\
					
					& 11  & Votes              & 435        & 16     & 1.60       & Yes      
					& 27  & Spambase           & 4,601      & 57     & 0.65       & No 	\\
					
					& 12  & Musk-clean-1       & 476        & 166    & 0.77       & No       
					& 28  & Mushroom           & 5,644(8,124) & 22   & 1.62       & Attr\#11 		 	\\
					
					& 13  & Wdbc               & 569        & 30     & 1.69       & No       
					& 29  & Cod-rna            & 59,535     & 8      & 0.50       & No 			\\
					
					& 14  & Credit-app         & 653(690)   & 15     & 1.21       & 37       
					& 30  & Ijcnn1             & 141,691    & 22     & 0.11       & No 			\\
					
					& 15  & Breast-cancer-W    & 683(699)   & 9      & 0.54       & 16 			
					& 31  & Skin-nonskin       & 245,057    & 3      & 0.26       & No 			\\
					
					& 16  & Statlog-australian & 690        & 14     & 0.81       & Yes 			
					&     &      &     &      &      &   			\\ \hline
				\end{tabular} 
			\end{sc}
		\end{small}
	\end{center}
	\vskip -0.1in
\end{table*}

\begin{table*}[t!]
	
	\caption{Comparison of average accuracies with its standard deviations and ranks.} \label{tbl.avgacc}
	
	\vskip 0.15in
	\begin{center}
		\begin{small}
			\begin{sc}
				\fontsize{6.5}{8.5}\selectfont
				\centering 
				\begin{tabular}{l@{\hskip0.5pt}l@{\hskip2pt}c@{\hskip2pt}c@{\hskip2pt}c@{\hskip2pt}c@{\hskip2pt}c@{\hskip2pt}c@{\hskip2pt}}
					\hline
					No. {\rule{0pt}{3ex}} & Data sets          & PE                        & PA                        & PATER-I                    & PATER-II                    & wPATER-I                   & wPATER-II                   \\ \hline
					1   {\rule{0pt}{3ex}} & Monks-3            & 70.902 $\pm$ 6.446 (6)    & 71.066 $\pm$ 4.795 (5)    & 75.574 $\pm$ 4.218 (4)    & 75.738 $\pm$ 3.821 (3)    & 75.984 $\pm$ 4.297 (2)    & \textbf{77.295 $\pm$ 4.675 (1)}    \\
					2   & Monks-1            & 64.355 $\pm$ 6.407 (5)    & 64.355 $\pm$ 7.760 (6)    & 64.919 $\pm$ 7.454 (4)    & 67.177 $\pm$ 3.747 (2)    & 67.097 $\pm$ 5.976 (3)    & \textbf{69.919 $\pm$ 3.891 (1)}    \\
					3   & Monks-2            & 54.195 $\pm$ 6.741 (3)    & 52.426 $\pm$ 6.664 (4)    & 49.235 $\pm$ 4.245 (6)    & 50.049 $\pm$ 5.094 (5)    & 61.720 $\pm$ 4.682 (2)    & \textbf{62.132 $\pm$ 4.186 (1)}    \\
					4   & Wpbc               & 67.835 $\pm$ 5.274 (3)    & 65.670 $\pm$ 5.558 (4)    & 59.021 $\pm$ 3.899 (5)    & 55.309 $\pm$ 5.375 (6)    & \textbf{72.113 $\pm$ 4.443 (1)}    & 69.278 $\pm$ 4.749 (2)    \\
					5   & Parkinsons         & 76.563 $\pm$ 5.060 (3)    & 78.150 $\pm$ 6.388 (2)    & 67.586 $\pm$ 3.124 (5)    & 66.408 $\pm$ 2.809 (6)    & \textbf{82.618 $\pm$ 4.467 (1)}    & 74.822 $\pm$ 5.296 (4)    \\
					6   & Sonar              & 69.904 $\pm$ 5.202 (5)    & 72.067 $\pm$ 3.424 (4)    & 72.788 $\pm$ 4.335 (2)    & 69.615 $\pm$ 5.242 (6)    & \textbf{74.471 $\pm$ 3.618 (1)}    & 72.115 $\pm$ 6.378 (3)    \\
					7   & SPECTF-heart       & 72.361 $\pm$ 3.659 (3)    & 72.473 $\pm$ 4.124 (2)    & 55.580 $\pm$ 3.530 (5)    & 53.633 $\pm$ 2.681 (6)    & \textbf{76.405 $\pm$ 3.015 (1)}    & 70.447 $\pm$ 3.898 (4)    \\
					8   & Statlog-heart      & 78.667 $\pm$ 4.303 (5)    & 77.667 $\pm$ 5.370 (6)    & 81.000 $\pm$ 2.867 (4)    & 82.815 $\pm$ 2.704 (2)    & 82.111 $\pm$ 2.956 (3)    & \textbf{84.037 $\pm$ 2.046 (1)}    \\
					9   & BUPA-liver         & 58.233 $\pm$ 4.359 (4)    & 57.769 $\pm$ 3.575 (5)    & 59.535 $\pm$ 2.958 (2)    & 54.463 $\pm$ 3.414 (6)    & \textbf{61.161 $\pm$ 3.785 (1)}    & 58.550 $\pm$ 4.726 (3)    \\
					10  & Ionosphere         & 80.573 $\pm$ 4.352 (5)    & 84.302 $\pm$ 2.641 (2)    & 81.711 $\pm$ 5.333 (3)    & 73.478 $\pm$ 6.391 (6)    & \textbf{85.128 $\pm$ 3.707 (1)}    & 81.367 $\pm$ 4.494 (4)    \\
					11  & Votes              & 90.481 $\pm$ 3.674 (3)    & 91.605 $\pm$ 3.796 (2)    & 90.369 $\pm$ 1.994 (4)    & 87.999 $\pm$ 1.633 (6)    & \textbf{92.368 $\pm$ 1.608 (1)}    & 88.644 $\pm$ 1.396 (5)    \\
					12  & Musk-clearn-1      & 70.399 $\pm$ 5.390 (4)    & 70.546 $\pm$ 5.285 (3)    & 74.958 $\pm$ 3.450 (2)    & 62.836 $\pm$ 6.602 (6)    & \textbf{75.546 $\pm$ 3.032 (1)}    & 63.592 $\pm$ 5.455 (5)    \\
					13  & Wdbc               & 94.728 $\pm$ 2.591 (4)    & 95.518 $\pm$ 1.985 (3)    & 96.010 $\pm$ 0.963 (2)    & 93.145 $\pm$ 1.369 (6)    & \textbf{96.116 $\pm$ 0.886 (1)}    & 93.479 $\pm$ 1.225 (5)    \\
					14  & Credit-app         & 80.503 $\pm$ 5.775 (6)    & 81.682 $\pm$ 5.579 (5)    & 84.349 $\pm$ 1.957 (4)    & 85.284 $\pm$ 1.519 (2)    & 84.441 $\pm$ 1.818 (3)    & \textbf{85.376 $\pm$ 1.581 (1)}    \\
					15  & Breast-cancer-W    & 95.623 $\pm$ 2.029 (5)    & 95.491 $\pm$ 2.126 (6)    & 96.969 $\pm$ 0.858 (3)    & 96.881 $\pm$ 0.581 (4)    & 97.115 $\pm$ 0.792 (2)    & \textbf{97.438 $\pm$ 0.593 (1)}    \\
					16  & Statlog-australian & 78.841 $\pm$ 5.582 (6)    & 79.493 $\pm$ 5.908 (5)    & 84.217 $\pm$ 2.009 (4)    & 85.507 $\pm$ 1.128 (2)    & 84.290 $\pm$ 1.702 (3)    & \textbf{85.580 $\pm$ 1.289 (1)}    \\
					17  & Blood-transfusion  & 68.436 $\pm$ 8.363 (3)    & 66.444 $\pm$ 11.36 (4)   & 59.024 $\pm$ 9.134 (6)    & 62.794 $\pm$ 3.059 (5)    & 76.377 $\pm$ 1.439 (2)    & \textbf{76.912 $\pm$ 1.255 (1)}    \\
					18  & Pima-diabetes      & 68.099 $\pm$ 3.994 (5)    & 69.505 $\pm$ 3.482 (4)    & 66.797 $\pm$ 5.272 (6)    & 72.057 $\pm$ 1.466 (2)    & 70.208 $\pm$ 3.635 (3)    & \textbf{74.740 $\pm$ 1.262 (1)}    \\
					19  & Mammographic       & 70.482 $\pm$ 16.87 (6)   & 72.241 $\pm$ 13.58 (5)   & 76.494 $\pm$ 8.903 (4)    & 81.349 $\pm$ 1.096 (2)    & 78.012 $\pm$ 6.420 (3)    & \textbf{81.602 $\pm$ 1.023 (1)}    \\
					20  & Tic-tac-toe        & 55.929 $\pm$ 5.301 (5)    & 56.013 $\pm$ 5.805 (4)    & 52.453 $\pm$ 4.740 (6)    & 56.983 $\pm$ 2.743 (3)    & 65.376 $\pm$ 1.731 (2)    & \textbf{67.056 $\pm$ 1.786 (1)}    \\
					21  & Statlog-german     & 68.470 $\pm$ 3.223 (4)    & 66.410 $\pm$ 4.967 (5)    & 63.120 $\pm$ 2.671 (6)    & 68.630 $\pm$ 1.454 (3)    & 71.910 $\pm$ 1.475 (2)    & \textbf{75.630 $\pm$ 1.260 (1)}    \\
					22  & Ozone-eight        & 89.924 $\pm$ 3.083 (3)    & 90.921 $\pm$ 3.164 (2)    & 52.528 $\pm$ 1.977 (6)    & 53.059 $\pm$ 1.154 (5)    & \textbf{92.534 $\pm$ 0.891 (1)}    & 67.845 $\pm$ 5.597 (4)    \\
					23  & Ozone-one          & 92.516 $\pm$ 5.130 (3)    & 95.070 $\pm$ 3.190 (2)    & 50.168 $\pm$ 1.519 (5)    & 49.800 $\pm$ 0.945 (6)    & \textbf{96.374 $\pm$ 0.478 (1)}    & 79.015 $\pm$ 10.88 (4)   \\
					24  & 20News-talk        & 49.957 $\pm$ 1.196 (4)    & 49.854 $\pm$ 1.030 (5)    & 49.811 $\pm$ 1.829 (6)    & 50.060 $\pm$ 0.986 (3)    & \textbf{51.292 $\pm$ 2.147 (1)}    & 50.838 $\pm$ 1.230 (2)    \\
					25  & 20News-comp        & 54.544 $\pm$ 4.127 (5)    & 53.376 $\pm$ 5.999 (6)    & 58.839 $\pm$ 6.779 (3)    & 57.495 $\pm$ 6.426 (4)    & \textbf{63.676 $\pm$ 6.350 (1)}    & 59.172 $\pm$ 9.193 (2)    \\
					26  & 20News-sci         & 72.572 $\pm$ 10.24 (4)   & 69.980 $\pm$ 9.689 (5)    & 73.310 $\pm$ 9.626 (3)    & 65.591 $\pm$ 5.703 (6)    & 74.420 $\pm$ 10.16 (2)   & \textbf{74.425 $\pm$ 6.073 (1)}    \\
					27  & Spambase           & 87.142 $\pm$ 2.570 (5)    & 86.707 $\pm$ 2.714 (6)    & 89.961 $\pm$ 1.743 (2)    & 89.048 $\pm$ 0.462 (4)    & \textbf{90.915 $\pm$ 0.894 (1)}    & 89.641 $\pm$ 0.411 (3)    \\
					28  & Mushroom           & 91.357 $\pm$ 5.435 (4)    & 93.698 $\pm$ 3.097 (3)    & 93.820 $\pm$ 1.721 (2)    & 87.509 $\pm$ 0.593 (6)    & \textbf{94.341 $\pm$ 0.891 (1)}    & 88.521 $\pm$ 0.761 (5)    \\
					29  & Cod-rna            & 90.590 $\pm$ 1.627 (2)    & \textbf{90.669 $\pm$ 1.822 (1)}    & 87.965 $\pm$ 1.843 (4)    & 69.606 $\pm$ 0.246 (6)    & 89.420 $\pm$ 2.441 (3)    & 77.571 $\pm$ 0.110 (5)    \\
					30  & Ijcnn1             & 89.655 $\pm$ 1.811 (4)    & 90.342 $\pm$ 1.673 (3)    & 59.430 $\pm$ 3.303 (6)    & 65.847 $\pm$ 0.690 (5)    & \textbf{92.029 $\pm$ 0.266 (1)}    & 90.426 $\pm$ 0.070 (2)    \\
					31  & Skin-nonskin       & 88.917 $\pm$ 5.564 (3)    & \textbf{91.104 $\pm$ 4.185 (1)}    & 89.302 $\pm$ 1.610 (2)    & 66.539 $\pm$ 4.338 (6)    & 88.209 $\pm$ 0.790 (4)    & 87.725 $\pm$ 0.077 (5)    \\
					& Average {\rule{0pt}{3ex}} & 75.573 $\pm$ 5.012 (4.19) & 75.891 $\pm$ 4.863 (3.87) & 71.511 $\pm$ 3.737 (4.06) & 69.571 $\pm$ 2.757 (4.52) & \textbf{79.477 $\pm$ 2.929 (1.77)} & 76.619 $\pm$ 3.125 (2.58) \\ \hline
				\end{tabular}
			\end{sc}
		\end{small}
	\end{center}
	\vskip -0.1in
\end{table*}

In order to establish wPATER-I and wPATER-II, similar to \eqref{eq.PATER1} and \eqref{eq.PATER2}, we get:
\begin{equation} \label{eq.wPATER1}
%\small
\begin{aligned}
\text{wPATER-I: }\mathcal{L}( \tau ) & =-\frac{1}{2} \tau ^{2}\left\Vert \mathbf{z}_{t}\right\Vert ^{2} +\tau \left( \frac{\alpha^- \lambda ^{-}_{t}}{n^{-}_{t}}\left( 1+\pmb{w}_{t}\mathbf{\cdot x}^{}_{t}\right)\right. \\
&\phantom{=} \left. +\frac{\alpha^+ \lambda ^{+}_{t}}{n^{+}_{t}}\left( 1-\pmb{w}_{t}\mathbf{\cdot x}^{}_{t}\right)\right) ,
\end{aligned}
\end{equation}
\begin{equation} \label{eq.wPATER2}
%\small
\begin{aligned}
\text{wPATER-II: }\mathcal{L}( \tau ) &  = -\frac{1}{2} \tau ^{2}\left\Vert \mathbf{z}_{t}\right\Vert ^{2} +\tau \left(\alpha^- k^{-}_{t} +  \alpha^+ k^{+}_{t}\right) .
\end{aligned}
\end{equation}
Taking the derivative of $\mathcal{L}( \tau )$ in \eqref{eq.wPATER1} and \eqref{eq.wPATER2} with respect to $\tau$ and setting it zero as:
\begin{equation}  
%\small
\begin{aligned}
\text{wPATER-I: }\frac{\partial \mathcal{L}( \tau )}{\partial \tau } & =-\tau \left\Vert \mathbf{z}_{t}\right\Vert ^{2} +\frac{\alpha^- \lambda ^{-}_{t}}{n^{-}_{t}}\left( 1+\pmb{w}_{t}\mathbf{\cdot x}^{}_{t}\right)\\
&\phantom{=} +\frac{\alpha^+ \lambda ^{+}_{t}}{n^{+}_{t}}\left( 1-\pmb{w}_{t}\mathbf{\cdot x}^{}_{t}\right)=0 ,
\end{aligned}
\end{equation}
\begin{equation} 
%\small
\text{wPATER-II: }\frac{\partial \mathcal{L}( \tau )}{\partial \tau } =-\tau \left\Vert \mathbf{z}_{t}\right\Vert ^{2} + \alpha^- k^{-}_{t} +\alpha^+ k^{+}_{t}=0 ,
\end{equation}
we get:
\begin{equation} \label{eq.wPATER1_tau}
%\small
\text{wPATER-I: } \tau _{t} =\frac{\frac{\alpha^- \lambda ^{-}_{t}}{n^{-}_{t}}\left( 1+\pmb{w}_{t}\mathbf{\cdot x}^{}_{t}\right) +\frac{\alpha^+ \lambda ^{+}_{t}}{n^{+}_{t}}\left( 1-\pmb{w}_{t}\mathbf{\cdot x}^{}_{t}\right)}{\left\Vert \mathbf{z}_{t}\right\Vert ^{2}} ,
\end{equation}
\begin{equation} \label{eq.wPATER2_tau}
%\small
\text{wPATER-II: } \tau _{t} =\frac{\alpha^- k^{-}_{t} + \alpha^+ k^{+}_{t}}{\left\Vert \mathbf{z}_{t}\right\Vert ^{2}} .
\end{equation}

\begin{figure*}[ht]
	\vskip 0.2in
	\begin{center}
		\subfigure[20News-sci]{ 
			\includegraphics[width=0.31\textwidth] 	{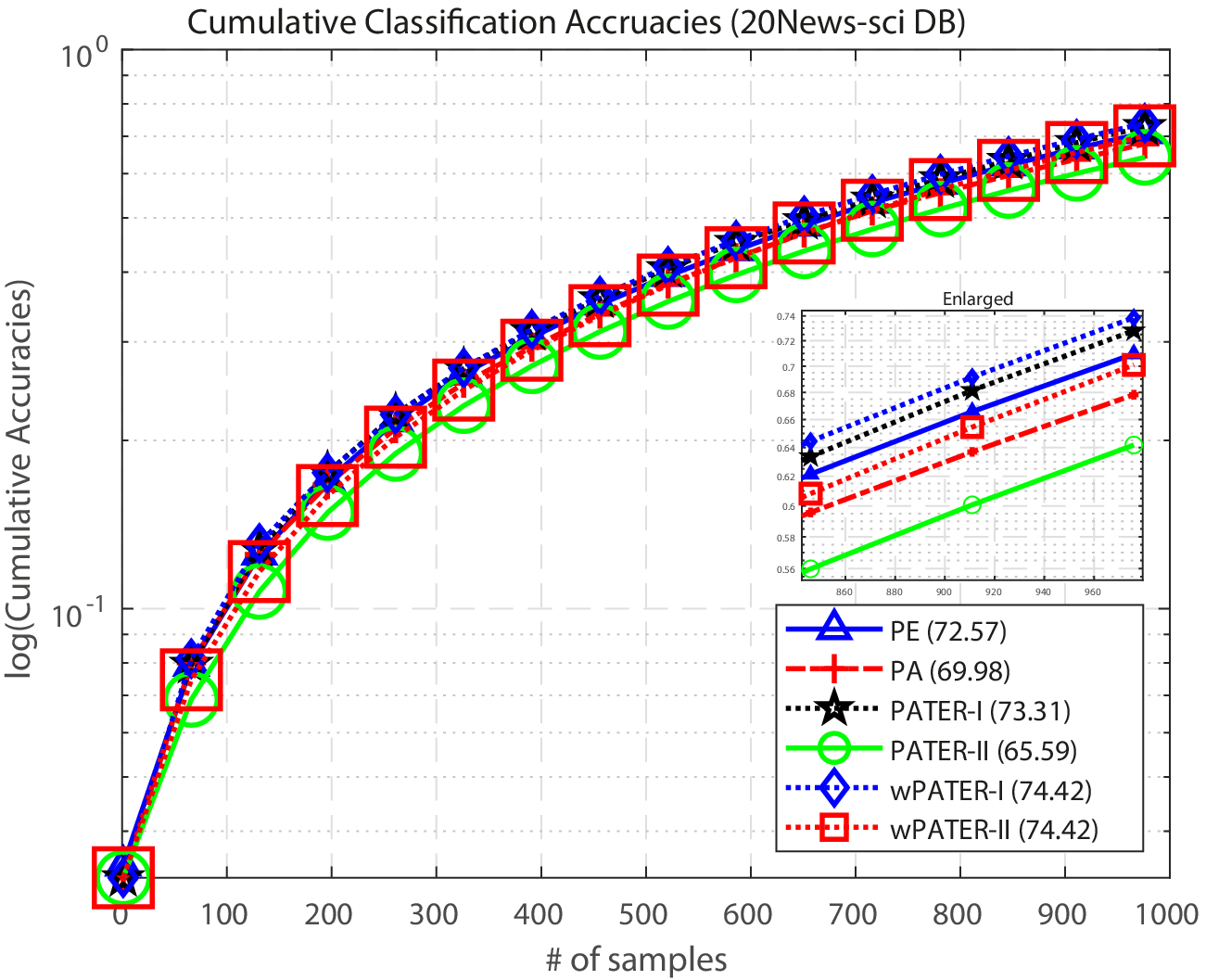}% 
		}  
		\subfigure[Spambase]{ 
			\includegraphics[width=0.31\textwidth] 	{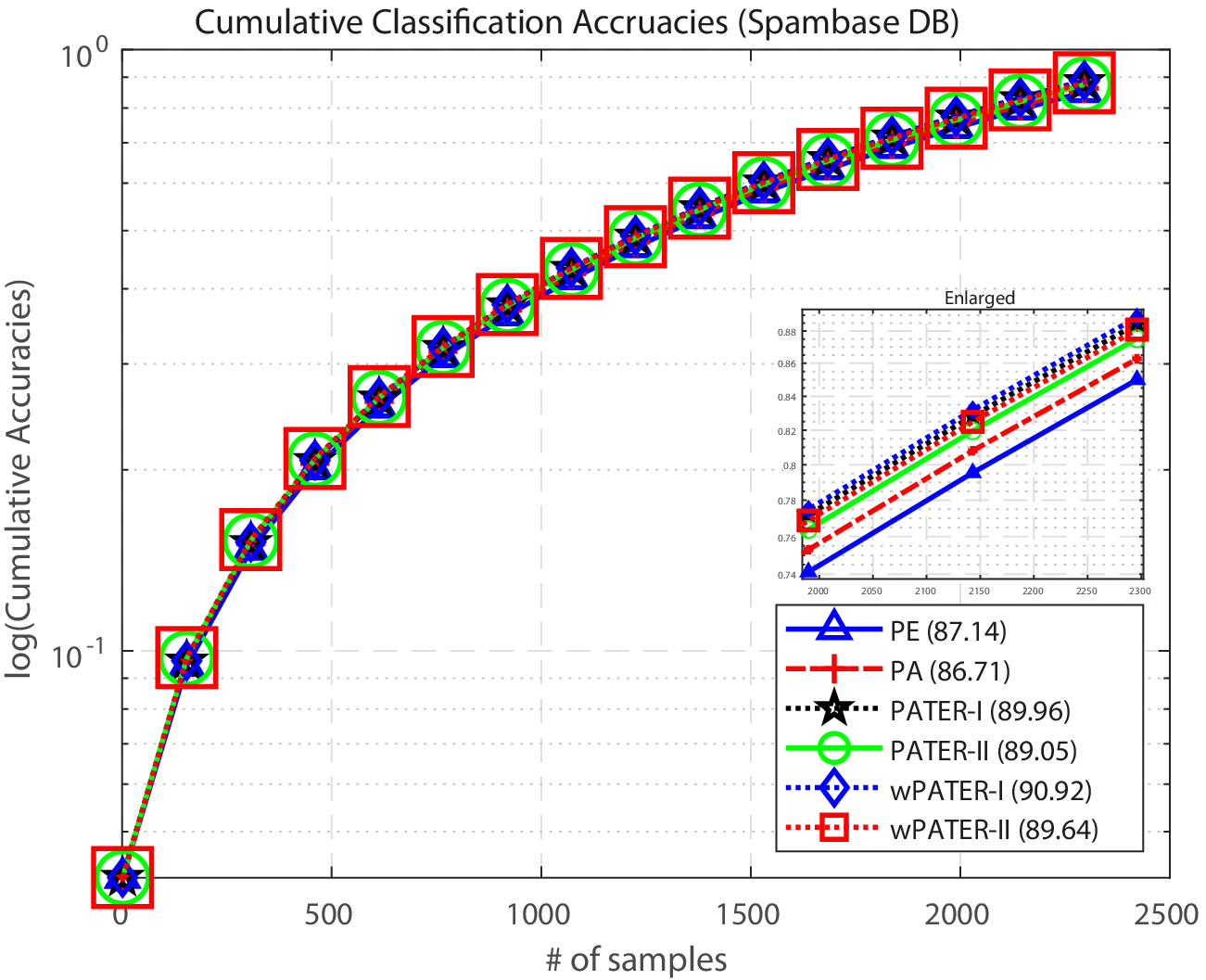}% 
		} 
		\subfigure[Mushroom]{ 
			\includegraphics[width=0.31\textwidth]  {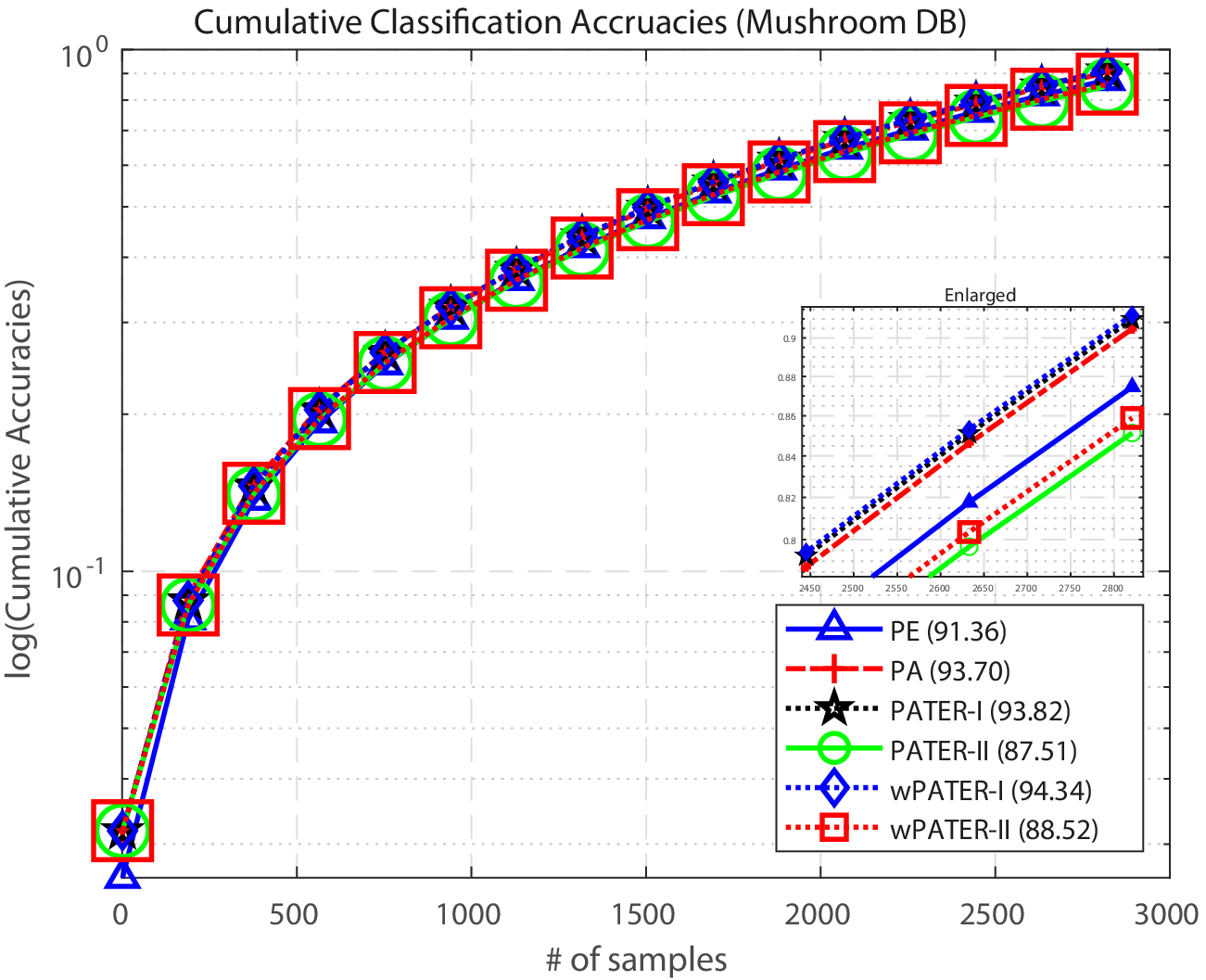}% 
		} 
		\subfigure[Cod-rna]{ 
			\includegraphics[width=0.31\textwidth] 	{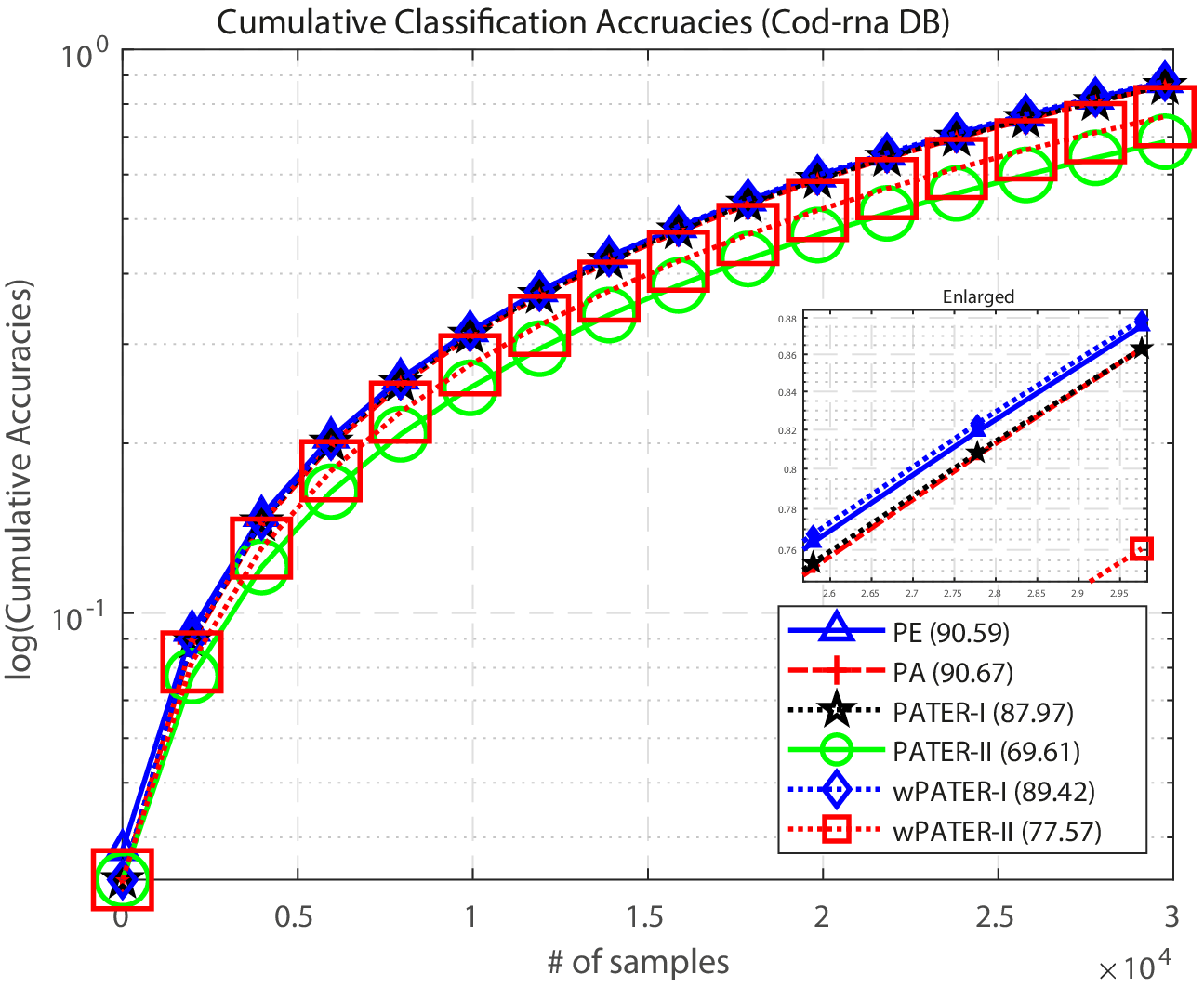}% 
		}  
		\subfigure[Ijcnn1]{ 
			\includegraphics[width=0.31\textwidth] 	{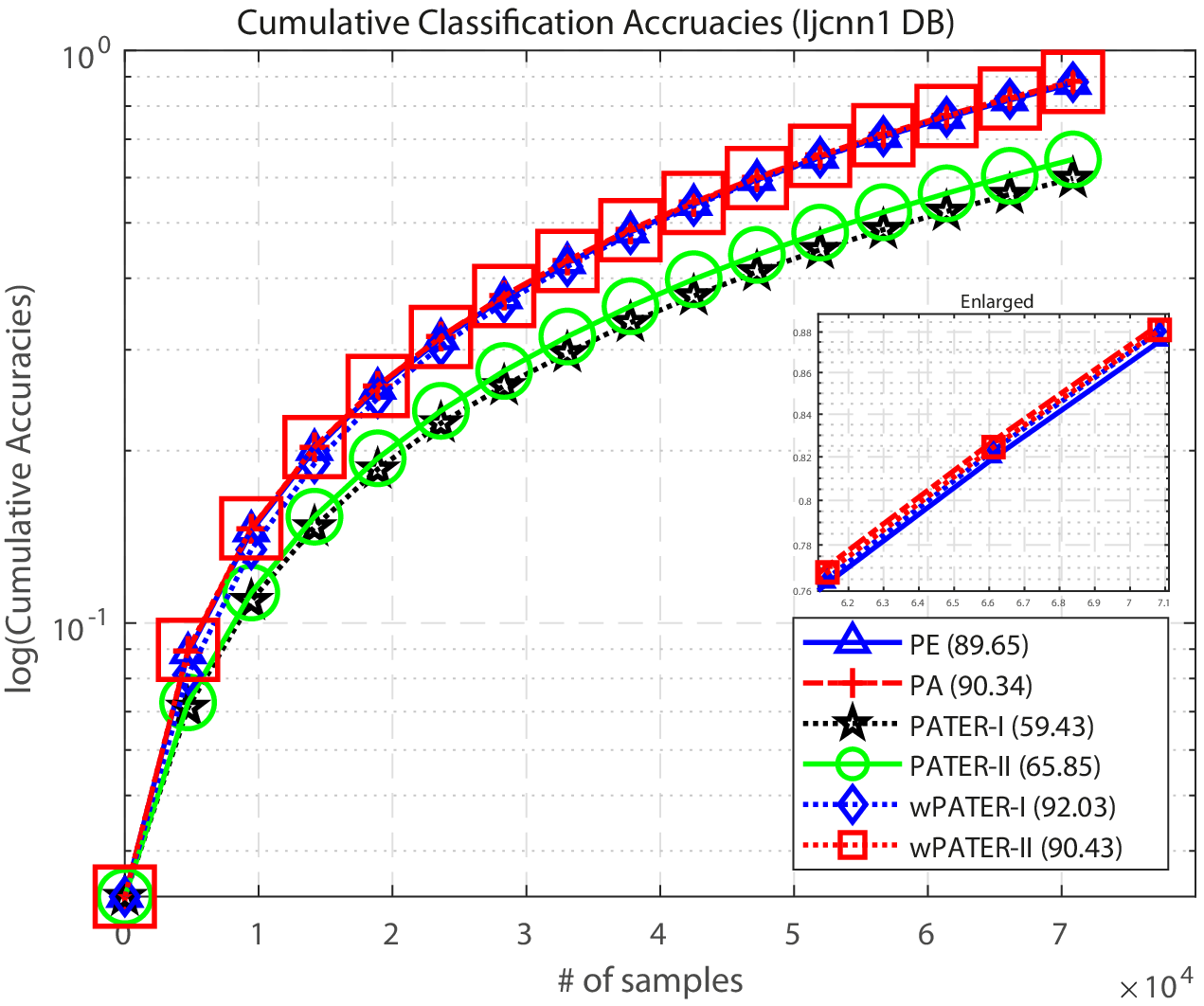}% 
		} 
		\subfigure[Skin-nonskin]{ 
			\includegraphics[width=0.31\textwidth] 	{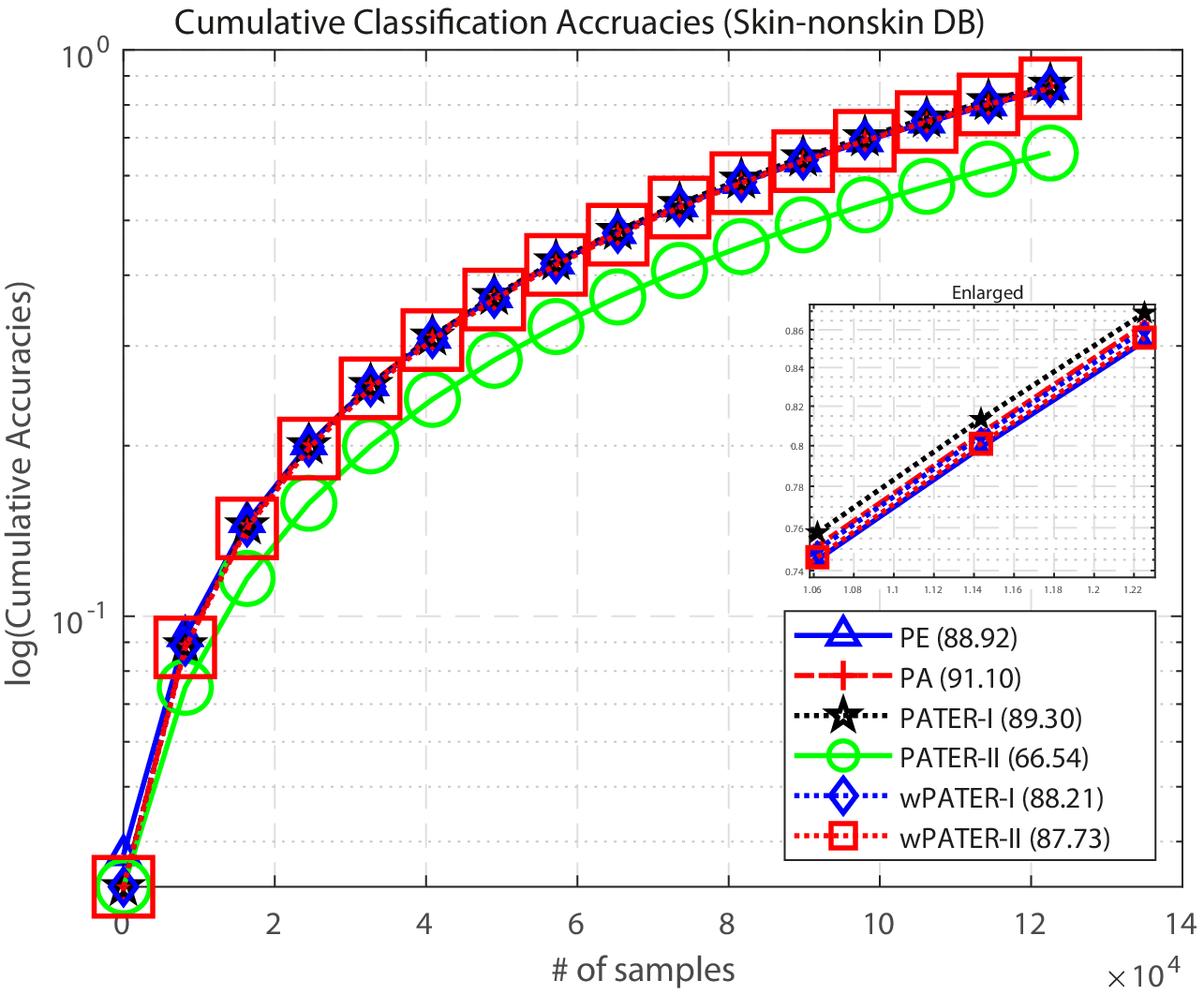}% 
		} 
		%		\centerline{\includegraphics[width=\columnwidth]{acc_Nemenyi_PATER}}
		\caption{The cumulative classification accuracies along the online learning process. The enlarged figure is also drawn for detailed comparison. The brackets include the average accuracies.}
		\label{fig.cma}
	\end{center}
	\vskip -0.2in
\end{figure*}

\subsection{Summary}
The pseudo code of the proposed wPATER algorithm is given in Algorithm \ref{alg.wPATER1}.
The main difference from PATER is that 
wPATER has two weights, $\alpha^-$ and $\alpha^+$, as hyperparameters in its loss function \eqref{eq.l_hinge_wTER} to address the imbalance data issue.

\section{Experiments}
In this section, we evaluate the performance of the proposed PATER and wPATER for binary classification. 
In terms of efficiency and effectiveness, the performance comparison is presented based on 31 real-world data sets.

\subsection{Data sets and experimental settings}
In our evaluation, 31 real-world data sets are obtained from 
UCI machine learning repository\footnote{\url{https://archive.ics.uci.edu/ml/index.php}} \cite{Lichman2013}, 
LIBSVM\footnote{\url{https://www.csie.ntu.edu.tw/~cjlin/libsvm/}} \cite{chang2011libsvm}
and 20 Newsgroups\footnote{\url{http://qwone.com/~jason/20Newsgroups/}} which are popular in the NLP community.
The description of the 31 data sets is summarized in Table \ref{tbl.dataSummary}.
Each data set is conducted in a z-score normalization.
The data set size ranges from 122 to 245,057 samples.
The data imbalance ratio is calculated by $\frac{n^+}{n^-}$.

For comparison with competing state-of-the-arts, the following algorithms are used as baselines: the perceptron (PE) \cite{rosenblatt1958perceptron}  and 
the online passive-aggressive algorithm (PA) \cite{crammer2006online}.
The average accuracies and CPU times are recorded using 10 runs of 2-fold cross-validation for each algorithm.

In wPATER, the two weights are varied as $\alpha^-, \alpha^+ \in \{0.01,\ 0.1,\ 0.3,\ 0.5,\ 0.9,\ 0.99\}$. When one weight is changed, another weight is fixed at $1$ (i.e., $\alpha^-=0.1$, $\alpha^+ =1$) to only address one side population of the imbalance binary classification problems.

\begin{table*}[t]
	
	\caption{Evaluation of the best weights between $\alpha^-$ and $\alpha^+ $ of the proposed wPATER  for the data imbalance problems. } \label{tbl.eval_weights}
	
	\vskip 0.15in
	\begin{center}
		\begin{small}
			\begin{sc}
				\fontsize{7}{8.5}\selectfont
				\centering 
				
				\begin{tabular}{l c@{\hskip4.5pt}c@{\hskip4.5pt}c@{\hskip4.5pt}c@{\hskip4.5pt}c@{\hskip4.5pt}c@{\hskip4.5pt}c@{\hskip4.5pt}c@{\hskip4.5pt}c@{\hskip3pt}c@{\hskip3pt}c@{\hskip3pt}c@{\hskip3pt}c@{\hskip3pt}c@{\hskip3pt}c@{\hskip3pt}c@{\hskip3pt}c@{\hskip3pt}c@{\hskip3pt}c@{\hskip3pt}c@{\hskip3pt}c@{\hskip3pt}c@{\hskip3pt}c@{\hskip3pt}c@{\hskip3pt}c@{\hskip3pt}c@{\hskip3pt}c@{\hskip3pt}c@{\hskip3pt}c@{\hskip3pt}c@{\hskip3pt}c@{\hskip3pt}c@{\hskip3pt}}
					\hline
					{\rule{0pt}{3ex}} & \multicolumn{31}{l}{Evaluation}                                                                                                                 & Sum \\ \hline

					Data No.  {\rule{0pt}{3ex}} & 1 & 2 & 3 & 4 & 5 & 6 & 7 & 8 & 9 & 10 & 11 & 12 & 13 & 14 & 15 & 16 & 17 & 18 & 19 & 20 & 21 & 22 & 23 & 24 & 25 & 26 & 27 & 28 & 29 & 30 & 31 &     \\
					Needed Weight (NW)     & P & N & P & P & N & N & N & P & N & P  & N  & P  & N  & N  & P  & P  & P  & P  & P  & P  & N  & P  & P  & N  & P  & N  & P  & N  & P  & P  & P  &     \\
					Best Weight (BW)       & P & P & P & P & N & N & N & P & N & P  & N  & P  & N  & P  & N  & P  & P  & P  & P  & P  & N  & P  & P  & N  & P  & P  & P  & N  & -  & P  & -  &     \\
					Results (BW equals NW) & 1 & 0 & 1 & 1 & 1 & 1 & 1 & 1 & 1 & 1  & 1  & 1  & 1  & 0  & 0  & 1  & 1  & 1  & 1  & 1  & 1  & 1  & 1  & 1  & 1  & 0  & 1  & 1  & 0  & 1  & 0  & 25  \\ \hline
				\end{tabular}
				\begin{flushleft}
					Notes: `N' and `P' indicate the negative and posivie classes respectively. 
					In NW, `N' is given if $\frac{n^+}{n^-} \geq 1$; otherwise `P' is given. 
					$\frac{n^+}{n^-} $ is the data imbalance ratio shown in Table \ref{tbl.dataSummary}. 
					In BW, `N' is given if the best performance is obtained from the $\alpha^- $ variation; otherwise `P' is given.
					%						NW is given by the ratio for each data set, where `N' is given if ${1\over1}$
				\end{flushleft}
			\end{sc}
		\end{small}
	\end{center}
	\vskip -0.1in
\end{table*}

\begin{table}[t]
	
	\caption{Comparison of average CPU times in seconds.} \label{tbl.avgtime}
	
	\vskip 0.15in
	\begin{center}
		\begin{small}
			\begin{sc}
				\fontsize{6}{8.5}\selectfont
				\centering 
				
				\begin{tabular}{l@{\hskip0.5pt}l@{\hskip2pt}c@{\hskip3.5pt}c@{\hskip3.5pt}c@{\hskip2.5pt}c@{\hskip2.5pt}c@{\hskip2.5pt}c@{\hskip2.5pt}}
					\hline
					No. {\rule{0pt}{3ex}}& Data sets          & PE     & PA     & PATER-I  & PATER-II & wPATER-I & wPATER-II \\ \hline
					1   {\rule{0pt}{3ex}}& Monks-3            & 0.0008 & 0.0013 & 0.0017 & 0.0019 & 0.0006  & 0.0009  \\
					2   & Monks-1            & 0.0006 & 0.0006 & 0.0008 & 0.0009 & 0.0006  & 0.0008  \\
					3   & Monks-2            & 0.0007 & 0.0008 & 0.0010 & 0.0012 & 0.0009  & 0.0012  \\
					4   & Wpbc               & 0.0006 & 0.0008 & 0.0011 & 0.0014 & 0.0010  & 0.0014  \\
					5   & Parkinsons         & 0.0006 & 0.0007 & 0.0010 & 0.0014 & 0.0009  & 0.0013  \\
					6   & Sonar              & 0.0007 & 0.0009 & 0.0013 & 0.0016 & 0.0011  & 0.0015  \\
					7   & SPECTF-heart       & 0.0008 & 0.0011 & 0.0015 & 0.0019 & 0.0014  & 0.0018  \\
					8   & Statlog-heart      & 0.0008 & 0.0010 & 0.0014 & 0.0019 & 0.0012  & 0.0018  \\
					9   & BUPA-liver         & 0.0011 & 0.0014 & 0.0019 & 0.0024 & 0.0016  & 0.0022  \\
					10  & Ionosphere         & 0.0010 & 0.0012 & 0.0019 & 0.0025 & 0.0017  & 0.0024  \\
					11  & Votes              & 0.0012 & 0.0013 & 0.0021 & 0.0031 & 0.0018  & 0.0029  \\
					12  & Musk-clearn-1      & 0.0016 & 0.0020 & 0.0028 & 0.0036 & 0.0025  & 0.0034  \\
					13  & Wdbc               & 0.0015 & 0.0015 & 0.0027 & 0.0041 & 0.0023  & 0.0038  \\
					14  & Credit-app         & 0.0019 & 0.0022 & 0.0031 & 0.0046 & 0.0028  & 0.0043  \\
					15  & Breast-cancer-W    & 0.0017 & 0.0017 & 0.0030 & 0.0048 & 0.0028  & 0.0045  \\
					16  & Statlog-australian & 0.0019 & 0.0023 & 0.0033 & 0.0048 & 0.0029  & 0.0046  \\
					17  & Blood-transfusion  & 0.0022 & 0.0026 & 0.0038 & 0.0051 & 0.0038  & 0.0049  \\
					18  & Pima-diabetes      & 0.0023 & 0.0028 & 0.0039 & 0.0054 & 0.0035  & 0.0051  \\
					19  & Mammographic       & 0.0024 & 0.0028 & 0.0039 & 0.0058 & 0.0037  & 0.0054  \\
					20  & Tic-tac-toe        & 0.0031 & 0.0038 & 0.0051 & 0.0065 & 0.0046  & 0.0063  \\
					21  & Statlog-german     & 0.0031 & 0.0038 & 0.0053 & 0.0070 & 0.0046  & 0.0066  \\
					22  & Ozone-eight        & 0.0056 & 0.0064 & 0.0110 & 0.0136 & 0.0086  & 0.0135  \\
					23  & Ozone-one          & 0.0054 & 0.0061 & 0.0109 & 0.0136 & 0.0084  & 0.0135  \\
					24  & 20News-talk        & 0.0061 & 0.0073 & 0.0096 & 0.0119 & 0.0088  & 0.0122  \\
					25  & 20News-comp        & 0.0062 & 0.0075 & 0.0097 & 0.0129 & 0.0090  & 0.0123  \\
					26  & 20News-sci         & 0.0058 & 0.0072 & 0.0093 & 0.0132 & 0.0085  & 0.0127  \\
					27  & Spambase           & 0.0152 & 0.0151 & 0.0238 & 0.0354 & 0.0200  & 0.0316  \\
					28  & Mushroom           & 0.0171 & 0.0169 & 0.0266 & 0.0415 & 0.0222  & 0.0376  \\
					29  & Cod-rna            & 0.4536 & 0.3357 & 0.4590 & 0.7083 & 0.2496  & 0.4397  \\
					30  & Ijcnn1             & 2.1150 & 1.1109 & 2.0940 & 2.2597 & 0.6436  & 1.9357  \\
					31  & Skin-nonskin       & 3.3478 & 1.9708 & 2.7245 & 3.9448 & 1.0431  & 1.9172  \\
					& Average   {\rule{0pt}{3ex}}   & 0.1938 & 0.1136 & 0.1752 & 0.2299 & 0.0667  & 0.1449 \\ \hline
				\end{tabular}
				
			\end{sc}
		\end{small}
	\end{center}
	\vskip -0.1in
\end{table}

\begin{figure}[ht]
	\vskip 0.2in
	\begin{center}
		\subfigure[Accuracy rank]{ 
			\includegraphics[width=0.45\textwidth]{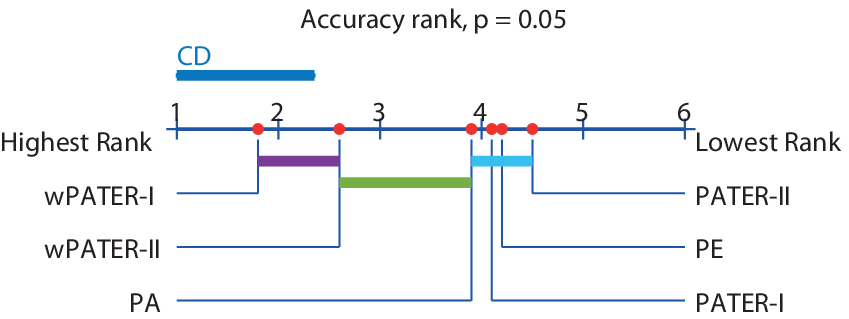}% 
		} \hspace{0.9cm}
		\subfigure[Computational rank]{ 
			\includegraphics[width=0.45\textwidth]{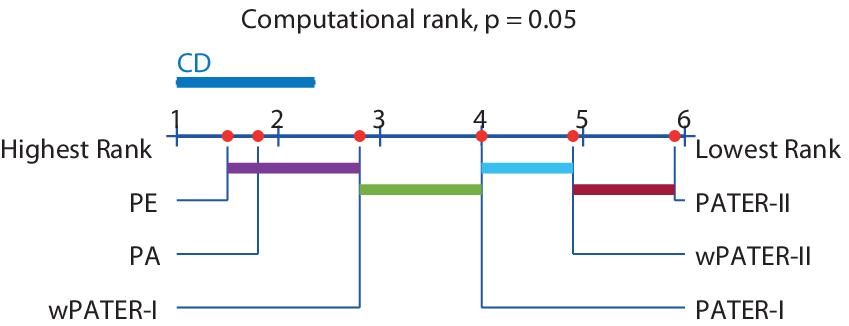}% 
		} 
		%		\centerline{\includegraphics[width=\columnwidth]{acc_Nemenyi_PATER}}
		\caption{The Nemenyi test to check statistical significance of the average (a) accuracies and (b) CPU times. The connected algorithms by the Critical Difference (CD) have no statistical significance.}
		\label{fig.Nemenyi}
	\end{center}
	\vskip -0.2in
\end{figure}

\subsection{Results and Discussion}
Table \ref{tbl.avgacc} shows the average accuracies and its standard deviations on the 31 real-world data sets.
Additionally, ranks for each algorithm are shown in brackets.
In terms of the average accuracy, the proposed wPATER-I performs better accuracy performance than that of the other algorithms. 
Moreover, the proposed wPATER-II is also shown as the second best.
In Table \ref{tbl.avgacc}, both wPATER-I and wPATER-II achieve the best performances on the 29 data sets. 
In order to investigate the validity  of the proposed wPATER for the data imbalance issue, 
in Table \ref{tbl.eval_weights}, we evaluate a best weight (BW), which gives the best accuracy, between $\alpha^-$ and $\alpha^+ $  for each data set. 
We then present matching results between the best weights and needed weights (NW) which are given by the data imbalance ratios from Table \ref{tbl.dataSummary}.
Consequently,  Table \ref{tbl.eval_weights} shows that 
our assumption of the proposed wPATER on addressing the data imbalance problems properly works in the 25 data sets.
Figure \ref{fig.cma} presents cumulative classification accuracies along the online learning process on the last 6 data sets (the data sets 26-31) that have large training samples.
%The remaining figures of the other data sets  can be found in the supplementary material.
Except the `Ijcnn1' data set in Figure \ref{fig.cma}(e), the PATER based algorithms maintain the best performers in terms of the cumulative accuracies.
%The remaining figures of the other data sets  can be found in the supplementary material.
For time complexity, Table \ref{tbl.avgtime} shows the average CPU times on the aforementioned 31 data sets.
In terms of the average CPU time, the proposed wPATER-I shows better CPU performance than that of the other algorithms.

To evaluate the statistical significance on the reported accuracy and CPU time comparisons, 
Friedman tests (see \cite{demvsar2006statistical}) and then Nemenyi plots are performed as a post-hoc test to statistically group connected algorithms at a confidence threshold (e.g., $p = 0.05$).
In terms of the accuracy rank, Fig. \ref{fig.Nemenyi}(a) has three groups of algorithms, namely 
1) wPATER-I---wPATER-II, 2) wPATER-II---PA and 3) PA---PATER-I---PE---PATER-II.
The proposed wPATER-I and wPATER-II are shown as the best group.
In terms of the computational rank, Fig. \ref{fig.Nemenyi}(b) shows four groups of algorithms, namely 
1) PE---PA---wPATER-I,
2) wPATER-I---PATER-I,
3) PATER-I---wPATER-II and
4) wPATER-II---PATER-II.
The proposed wPATER-I is included in the best group, although this is shown as the third best individually. In summary, the proposed wPATER-I is seen as the best algorithm in terms of  efficiency and effectiveness.

\section{Conclusion}
We have presented a passive-aggressive total-error-rate (PATER) minimization for online learning. 
Consequently, the proposed PATER had the useful properties of PA and TER together.
In addition, a weighted PATER method (wPATER) has also presented to address the data imbalance problems.
The proposed wPATER effectively and efficiently outperformed the state-of-the-art algorithms, including the proposed PATER  on the real-world data sets.
This work was based on the first-order information. We will improve this work to extend it to the second-order methods.

\bibliography{references.bib}
\bibliographystyle{icml2020}

\end{document}